**Abbreviated title**

Identifying individual cow groups associated with milk traits

**Title**

# Meta-clustering of milk mid-infrared spectra identifies dairy cow groups associated with negative energy balance in early lactation

**Authors**

T. Touil[1,2,3], É. R. Paquet[1,2,3*]

**Affiliations:**

[1]Département des sciences animales, Université Laval, Québec, QC, G1V 0A6, Canada

[2]Institut intelligence et données, Université Laval, Québec, QC, G1V 0A6, Canada

[3]Centre de recherche en données massives, Université Laval, Québec, QC, G1V 0A6, Canada

*Corresponding author: eric.paquet@fsaa.ulaval.ca

**Highlights:**

- Eight clustering pipelines—combining spectral filtering, dimensionality reduction, and clustering algorithms—were applied directly to milk MIR spectra.
- Despite methodological differences, the eight approaches converged on five consistent meta-clusters of early-lactation cows.
- Using the full spectrum, the classic, computationally efficient PCA k-means approach recaptured the meta-clusters found by more sophisticated, computationally intensive approaches.
- The five meta-clusters were strongly associated with days in milk (DIM).
- Three meta-clusters reflected a gradient of negative energy balance (NEB) severity—severe, moderate, and mild—during early lactation.
- Two meta-clusters likely represented cows recovering from NEB: one with rapid restoration, one in early recovery.

# Abstract

Clustering methods have been used to identify distinct groups of milk samples, cows, or herds. Fourier-transform infrared (FTIR) spectroscopy, particularly mid-infrared (MIR) spectroscopy, has been applied to individual cow milk samples to predict various milk traits. Applying clustering directly to MIR spectral data may reveal latent groups of cows associated with milk traits or health disorders and can help prevent these conditions or monitor at-risk animals. This study aimed to identify groups of individual dairy cows in early lactation directly from milk MIR spectra and to analyze their associations with milk traits. Using a dataset of 407,632 individual milk MIR records from 3,408 commercial farms, we combined (i) spectral filtering that selects informative wavenumbers, (ii) two dimensionality-reduction methods: principal component analysis (PCA) and an autoencoder, and (iii) two clustering algorithms: k-means and spectral clustering to yield eight different clustering approaches. We regrouped the assigned clusters into meta-clusters that encompassed the most similar ones identified by the eight approaches. Our results revealed five distinct meta-clusters of early-lactation individual dairy cows significantly associated with milk traits. Despite substantial differences, the eight approaches converged on the same five meta-clusters, and the classic, computationally efficient PCA-based k-means approach using the full spectrum recaptured clusters identified by more sophisticated, computationally intensive approaches. The five meta-clusters were strongly associated with DIM and appeared to reflect a gradient of negative energy balance (NEB) severity: severe, moderate, and possibly mild, while the remaining two likely represented cows recovering from NEB, one with rapid restoration of energy balance and one in early recovery.

# Introduction

Monitoring cow health and performance is challenging and incurs substantial costs for dairy producers. Developing tools to support this monitoring is essential—not only to reduce costs, but also to prevent metabolic, physiological, and nutritional disorders, promote animal welfare, and ensure high milk quality. Some researchers have focused on identifying groups of dairy cows or atypical mid-infrared (MIR) spectra associated with milk traits or other factors—physiological, environmental, managerial, or pathological—using Fourier-transform infrared (FTIR) milk data. For example, Spieß et al. (2021) combined clustering algorithms with untargeted models to group atypical MIR spectra, representing abnormal milk samples.

Other studies have focused on identifying groups of cows or herds. Grelet et al. (2019) proposed an approach to group dairy cows based on the concentrations of four individual blood biomarkers, which they attempted to predict from MIR milk spectra. De Koster et al. (2019) clustered dairy cows in early lactation using the concentrations of four blood components, with MIR spectra serving as biomarkers to construct predictive models for the identified clusters. Franceschini et al. (2022) grouped cows using a large milk recording dataset composed of 29 biomarkers, including 27 predicted from individual cow MIR milk spectra. Franceschini et al. (2025) extended this approach from individual cow milk to bulk tank milk (clusters of herds), applying clustering to 31 fatty acids or fatty acid (FA) groups predicted from MIR spectra.

While these studies offer valuable insights, none have attempted to identify clusters of cows directly from MIR spectral data derived from milk samples collected from many individual cows on commercial farms. MIR spectral data may reveal latent groups or hidden patterns that other data types cannot capture and that are associated with distinct milk trait profiles, reflecting different levels of cows' metabolic state. Therefore, the objectives of this study were to (1) evaluate and compare several clustering approaches applied directly to MIR spectra from individual dairy cows in early lactation, (2) assess the extent to which these approaches agreed on common cluster structures, and (3) analyze the associations between the resulting clusters and milk traits, using MIR spectra from thousands of cows on commercial farms.

# Materials and methods

### *Data Collection and Generation of the MIR Dataset and Milk Traits*

A total of 407,632 milk samples were collected from 291,034 dairy cows housed in 3,408 commercial dairy farms across Québec, Canada, between March 2019 and June 2022. Cow parity ranged from 1 to 14, with a median of 2. The farms represented diverse management practices, feeding strategies, and housing systems. Milk recording and analysis were performed by Lactanet (a DHI laboratory based in Sainte-Anne-de-Bellevue, QC, Canada), using six FTIR instruments (MilkoScan FT+, FOSS; and MilkoScan 7 RM, FOSS; Hilleroed, Denmark) to generate MIR spectral data. Each spectrum provided absorbance values for 1,060 wavenumbers (5,010–925 $cm^{-1}$).

Fat, FAs (*de novo*, short-chain, medium-chain, long-chain, mixed, preformed, saturated, monounsaturated, polyunsaturated, C14:0, C16:0, C18:0, C18:1, *trans*, free), protein, lactose, milk urea nitrogen (MUN), β-hydroxybutyrate (BHB), somatic cell count (SCC), and days in milk (DIM) were provided by Lactanet. All milk traits except SCC were measured using the FTIR instruments; SCC was measured using a CombiFoss 7 instrument via flow cytometry, and the $\log_{10}$-transformed SCC was used in subsequent analyses. Fat-based versions of the FAs ($FA_{denovo_F}$, $FA_{short\text{-}chain_F}$, $FA_{medium\text{-}chain_F}$, $FA_{long\text{-}chain_F}$, $FA_{mixed_F}$, $FA_{preformed_F}$, $FA_{saturated_F}$, $FA_{monounsaturated_F}$, $FA_{polyunsaturated_F}$, $FA_{C14:0_F}$, $FA_{C16:0_F}$, $FA_{C18:0_F}$, $FA_{C18:1_F}$, $FA_{trans_F}$, $FA_{free_F}$: grams of fatty acid per 100 grams of total FAs) were calculated because expressing FAs relative to total FAs results in higher numerical concentrations than expressing them relative to milk (grams of fatty acid per 100 grams of milk) and provides more biologically meaningful information (Hanuš et al., 2018).

***Data Pre-Processing***

From the 407,632 MIR spectra dataset, one sample per cow was randomly selected, resulting in 291,034 spectra. Consequently, the clustering analysis focused on inter-cow differences rather than intra-cow variability. Only spectra from the three instruments contributing most samples were retained to reduce instrumental heterogeneity, limit instrument-related effects, and prevent the identification of clusters reflecting instrument differences rather than biological variation, yielding 214,869 spectra. These were standardized following Bonfatti et al. (2017) to correct for temporal calibration shifts. The standardized dataset contained 1,060 wavenumbers, which were further Z-score standardized (mean = 0, SD = 1). A standardized Mahalanobis distance based on principal component analysis (PCA) was computed to detect spectral outliers, following Soyeurt et al. (2020). Three outliers were removed, leaving a dataset of 214,866 spectra. Regarding spectral filtering the reference (**R**) approach retained all 1,060 wavenumbers while the (**F16**) approach retained the 212 informative wavenumbers defined by Grelet et al. (2016).

***Dimensionality Reduction and Clustering Workflow***

A PCA and an autoencoder deep learning method were used for dimensionality reduction, implemented in Python (version 3.10.15) (Van Rossum & Drake, 2009) with the scikit-learn library (Pedregosa et al., 2011) and TensorFlow (Abadi et al., 2016), respectively.

Initially, both methods were applied without F16 filtering. For PCA, 20 uncorrelated principal components (PCs) were retained, explaining 95.15% of the spectral variability and reducing the dataset from 1,060 to 20 features. The explained variance threshold was set at 95% throughout the study to retain the main data structures while removing low-variance dimensions primarily associated with noise, thereby preserving most of the dataset's variability during dimensionality reduction (Jolliffe & Cadima, 2016).

K-means and spectral clustering (scikit-learn; Pedregosa et al., 2011) were then applied separately to the reduced dataset of 214,866 spectra. The number of clusters was determined through exploratory tuning. Several values of k were tested to assess their impact on the resulting group structure. The elbow method identified k = 4 as an appropriate compromise. Accordingly, k-means was performed with k = 4 clusters (n_clusters = 4) and k-means++ initialization (init = 'k-means++'). To ensure a fair comparison between methods, the same value of k = 4 was used for spectral clustering. This choice was intended to isolate the effect of the clustering algorithm itself while avoiding bias introduced by differing k values. Since our dataset is large and spectral clustering is memory-intensive so we were only able to run it on 40% of our dataset. Therefore, 40% of the dataset (85,946 spectra) was randomly subsampled, and spectral clustering was performed on this subset to obtain an initial partition. Cluster labels from this initial partition were then propagated to the full dataset using a k-nearest neighbors (k-NN; scikit-learn; Pedregosa et al., 2011) classifier with n_neighbors = 10. Symmetric normalized spectral clustering was performed with the following hyperparameters: n_clusters = 4, an affinity matrix based on the k-NN method (affinity = 'nearest_neighbors', n_neighbors = 10), a k-NN graph representation, and cluster assignment using QR decomposition (assign_labels = 'cluster_qr').

For the autoencoder applied on all wavenumbers, dimensionality was reduced from 1,060 to 19 features, explaining 95.09% of spectral variability. The architecture included an input layer (1,060 neurons), three encoding layers (500, 500, and 2,000 neurons; ReLU activation), a 19-unit bottleneck (linear activation), and a mirrored decoder (2,000, 500, and 500 neurons; ReLU and tanh activations) with a linear output layer. Training used the Adam optimizer (learning rate = $1\times10^{-5}$), mean squared error loss, early stopping, and model checkpointing based on validation loss, with a batch size of 128 and 300 epochs. Only the encoder was used for dimensionality reduction. K-means and spectral clustering were applied to the reduced dataset using the same procedure and hyperparameters as for PCA, except that spectral clustering was run on a 30% subsample (64,459 spectra). Subsampling proportions ranging from 90% to 30% were evaluated, and 30% was the largest proportion that could be handled by the spectral clustering algorithm.

Subsequently, both PCA and the autoencoder were applied with F16 filtering, which reduced the dataset from 1,060 to 212 wavenumbers. The PCA retained three PCs, explaining 96.51% of variability, reducing the dataset to three features. K-means and spectral clustering were applied using the same hyperparameters as when using all wavenumbers (R), but here spectral clustering was run on the full dataset. For the autoencoder applied on F16, dimensionality was reduced from 212 to 3 features, explaining 97.04% of variability, using the same architecture as before but with an input layer of 212 neurons and a 3-unit bottleneck. K-means and spectral clustering were then applied using the same procedure as PCA with F16.

In total, eight different clustering approaches were tested: R PCA k-means, R PCA spectral clustering, R autoencoder k-means, R autoencoder spectral clustering, F16 PCA k-means, F16 PCA spectral clustering, F16 autoencoder k-means, and F16 autoencoder spectral clustering.

***Assembly of Cluster Assignments and Milk Traits***

Cluster assignments for each cow were combined with the corresponding milk traits and DIM. Associations of the obtained clusters with breed and DIM were evaluated. To facilitate the interpretation of cluster–trait relationships while reducing heterogeneity associated with breed composition and lactation stage, subsequent analyses were restricted to the major breeds (Holstein, Ayrshire, and Jersey) and to early lactation (DIM 0–49), which contained most observations. This resulted in a final dataset comprising 179,815 samples and 30 variables (eight clustering approaches, 21 milk traits, and DIM).

***Meta-Clustering Procedure***

Each clustering approach produces a distinct partition of the cows that is not directly comparable across approaches. Therefore, a meta-clustering method was applied to identify common clusters the 8 different clustering approaches. To achieve this, similarities between cluster assignments were computed for each cow using the Jaccard index in R (version 4.2.2; R Core Team, 2022). The resulting similarity matrix was then grouped into five meta-clusters using Ward's hierarchical agglomerative clustering algorithm (ward.D2) with squared Euclidean distance, implemented in the pheatmap package (Kolde, 2025). The partition at k = 5 maximizes inter-cluster separation while minimizing within-cluster heterogeneity, resulting in meaningful meta-clusters representing consistent clustering patterns. Each cow was assigned to a meta-cluster if ≥ 50% of its cluster assignments corresponded to that meta-cluster. Notched boxplots were used to compare DIM distributions among meta-clusters. Milk traits were standardized (Z-score), and their median values per meta-cluster were visualized in a heatmap (pheatmap; Kolde, 2025), with values limited to the range –1 to 1 to enhance contrast among meta-clusters.

# Results and discussion

***Distinct Clusters from the Eight Clustering Approaches***

Applying the eight clustering approaches produced four clusters of cows for each method (Figure 1). Because the approaches differed in spectral filtering, dimensionality reduction, and clustering algorithm, they generated distinct partitions of the data. Consequently, the same cow could be assigned to different clusters depending on the approach used. For example, cluster assignments obtained with k-means differed when applied after PCA or autoencoder dimensionality reduction, and additional differences were observed between k-means and spectral clustering. Overall, each cow could receive up to eight cluster assignments, one from each approach. These partitions were not directly comparable, motivating the subsequent meta-analysis aimed at identifying common clusters across the different approaches.

***Meta-Clusters Interpretation***

Figure 2 shows the heatmap of similarities between cluster assignments, computed using the Jaccard index. Five meta-clusters were identified from the resulting similarity matrix, representing groups of cluster assignments that overlapped most frequently. Meta-cluster 1 was identified by all clustering approaches except one, suggesting its potential importance. Meta-clusters 2 and 3 were each identified by six of the eight clustering approaches, indicating similar relevance. The R PCA k-means (PCA.KM) approach identified most meta-clusters (3, 2, 1, and 4), including those shared across multiple methods, as well as a rarer meta-cluster (meta-cluster 4) that was not detected by most other approaches. Since the R PCA k-means approach is deterministic, efficient and recapitulated most meta-clusters it was considered the most appropriate method for clustering cows in this study.

Associations between the meta-clusters and various animal, environmental, and analytical factors, including milking system, parity, DIM, season, FTIR instrument, year, and milk traits, were investigated. Given the large number of animals within each meta-cluster, ranging from n = 2,639 for metaC4 to n = 60,577 for metaC5, DIM and all milk production traits showed significant associations with the identified meta-clusters (Figures 3 and 4).

***Interpreting Associations of Meta-Clusters with DIM***

After assigning each meta-cluster to a cow, we observed varying proportions across the meta-clusters; meta-cluster 4 had the lowest percentage of cows (n = 2,639, 1.49%) while meta cluster 5 had the highest (n = 60,577, 34.16%, Figure 3). We then analyzed the distribution of DIM across these meta-clusters and found significant associations ($P < 0.0001$). Comparing DIM distributions for early-lactation cows indicated that cows in meta-clusters 1, 2, and 4 were likely experiencing negative energy balance (NEB; Grummer, 2008; Herdt, 2000)—the primary cause of fat mobilization from adipose tissue (lipomobilization; Herdt, 2000; Weber et al., 2013) and losses in body condition score (BCS) and body weight (BW) (Ferguson, 1996; Grummer & Rastani, 2003)—which typically occurs during the first 3 weeks of lactation (Carvalho et al., 2014; Civiero et al., 2021; Drackley, 1999; Nigussie, 2018). Meta-clusters 3 and 5 represented groups of cows likely entering the recovery phase from NEB, which generally begins after 3 weeks in milk. On average, cows return to positive energy balance at 45 DIM, with a standard deviation of 21 d (Grummer & Rastani, 2003; Noordhuizen, 2014).

***Interpreting Associations of Meta-Clusters with Milk Traits***

Median values of standardized milk traits across meta-clusters are shown in Figure 4. Fat, $FA_{long\text{-}chain_F}$, $FA_{monounsaturated_F}$, $FA_{preformed_F}$, $FA_{C18:1_F}$, $FA_{C18:0_F}$, BHB, and $FA_{trans_F}$ were elevated in meta-clusters 1 and 4, indicating high or moderate milk fat content and concentrations of these mostly preformed FAs in milk fat—mobilized from adipose tissue during lipomobilization—and increased BHB from fat mobilization, consistent with NEB (Churakov et al., 2021; J. Gross, Dorland, et al., 2011). The elevated BHB in meta-cluster 4 suggests possible subclinical ketosis (Cascone et al., 2022).

Conversely, lactose, $FA_{medium\text{-}chain_F}$, $FA_{denovo_F}$, $FA_{C14:0_F}$, $FA_{short\text{-}chain_F}$, $FA_{mixed_F}$, $FA_{C16:0_F}$, $FA_{saturated_F}$ were moderately or greatly reduced, indicating lower milk or fat content of these *de novo* and mixed FAs—mostly synthesized in the mammary gland—and reflecting reduced mammary synthesis due to energy deficiency, suggesting a NEB origin (Churakov et al., 2021; J. Gross, Dorland, et al., 2011; Knegsel et al., 2005). A decrease in lactose also suggests NEB (Hamon et al., 2024). Cows in meta-cluster 1, with higher level of preformed FAs, low *de novo* FAs, moderate BHB, and no strong lactose depression, were likely experiencing moderate NEB (Churakov et al., 2021; J. Gross, Dorland, et al., 2011; Hamon et al., 2024). Cows in meta-cluster 4, with high preformed FAs, low *de novo* FAs, high BHB, and strong lactose depression, were likely experiencing severe NEB (Churakov et al., 2021; Hamon et al., 2024; Lei & Simões, 2021).

Among the remaining milk traits, protein content decreased slightly only in meta-cluster 1, consistent with NEB-related prioritization of energy for survival over milk synthesis (Civiero et al., 2021; J. Gross, van Dorland, et al., 2011; J. J. Gross & Bruckmaier, 2019). SCC increased slightly only in meta-cluster 4, suggesting subclinical mastitis or metabolic stress (Carvalho-Sombra et al., 2021). $FA_{polyunsaturated_F}$ content was moderately elevated in meta-clusters 1 and 4, possibly reflecting polyunsaturated FA-rich diets that provide additional calories during NEB (Castro et al., 2019), reduce metabolic stress—as also suggested by SCC patterns in meta-cluster 4—or support inflammatory adaptation (Bionaz et al., 2020).

Additionally, the increased contents of $FA_{polyunsaturated_F}$ and $FA_{trans_F}$ observed in meta-cluster 4 could be explained by alterations in biohydrogenation and by the production of specific *trans*-10 isomers (*trans*-10, *cis*-12 18:2 and *trans*-10 18:1), respectively. These specific *trans*-10 isomers inhibit *de novo* FA synthesis in the mammary gland, thereby explaining its decreased level. This process suggests that cows in meta-cluster 4 were potentially experiencing subclinical ruminal acidosis (SARA) (Huot et al., 2023, 2024; Touil et al., 2025).

Meta-cluster 2 showed high milk protein content, potentially reflecting low-producing cows with low milk volume (Morton et al., 2016) likely experiencing mild NEB (Martens, 2023). Meta-cluster 3 showed the lowest fat and protein contents, with low concentrations of preformed FAs and BHB, suggesting a short, mild NEB that was quickly resolved (Churakov et al., 2021). Meta-cluster 5 displayed only mild deviations from the average profile (e.g., slight decreases in C18:0 and C18:1), possibly indicating early recovery from NEB as energy balance was being restored (Churakov et al., 2021).

Comparing our results with Grelet et al. (2019), who identified three groups of cows representing healthy, moderately impacted, or imbalanced metabolic status, our five meta-clusters could be grouped similarly: meta-clusters 3 and 5 in the healthy group, meta-clusters 1 and 2 in the moderately impacted group, and meta-cluster 4 in the imbalanced group. Our results also aligned with De Koster et al. (2019), who distinguished cows with balanced or imbalanced metabolic profiles associated with energy balance. They reported higher energy balance in balanced cows (i.e., cows with an adequate energy status and no marked NEB; corresponding to our meta-clusters 3 and 5) and lower energy balance in imbalanced cows (corresponding to our meta-clusters 1, 2, and 4). Meta-cluster 4 also corresponded to cluster 4 in Franceschini et al. (2022), which represented cows with metabolic disorders induced by NEB up to ketosis. In addition, it aligned with one of the seven clusters identified by Franceschini et al. (2025), which showed high levels of long-chain FAs—a sign of NEB—although their clusters were defined at the herd level, whereas ours were based on individual cows. Our findings extend previous work by identifying meta-clusters of individual cows associated with distinct degrees of NEB and by revealing a meta-cluster potentially associated with SARA.

***Limitations***

This study has some limitations. First, only milk-derived traits were available. If blood metabolites such as NEFA, BHB, and blood glucose—direct biomarkers of NEB—had been measured, we could have confirmed the association between our meta-clusters and NEB levels. In addition, including data on dry matter intake (DMI), BCS, and BW of cows would have further strengthened our interpretations. Second, we restricted the analysis to early-lactation cows to avoid bias from stage of lactation. A dataset with balanced numbers of early-, mid-, and late-lactation cows could yield additional clusters and deeper insights. Future studies should include blood metabolites, zootechnical trait data, and other relevant variables, as well as more balanced datasets, to validate the identified meta-clusters, detect additional clusters, and provide broader insights into cow health monitoring.

## Conclusion

In conclusion, five meta-clusters of individual dairy cows in early lactation associated with milk traits were identified using eight different clustering approaches combining spectral filtering, dimensionality reduction, and clustering algorithms applied directly to MIR data. Among these, the PCA k-means approach using all wavenumbers (R) emerged as the most appropriate method for clustering cows, as it identified most of the meta-clusters, including those shared across methods and a rarer meta-cluster detected by only a few other approaches and that this method is deterministic and efficent. All five meta-clusters were strongly associated with DIM. Three meta-clusters represented cows likely experiencing NEB: one included cows with severe NEB, possibly with subclinical ketosis, metabolic stress, and SARA; one included cows with moderate NEB; and one included cows that may have experienced mild NEB. The remaining two meta-clusters represented cows likely recovering from NEB—one with a short, mild NEB that was quickly resolved, and one in early recovery. Although these interpretations are less precise than those based on blood metabolites and require validation in future studies, this study may support dairy producers in herd monitoring and decision-making, thereby promoting overall cow health and performance.

## Declaration of Generative AI and AI-assisted technologies in the writing process

ChatGPT was used to improve the English, and all authors reviewed, verified, and edited the last version of this manuscript.

## Acknowledgments

The authors would like to thank Lactanet for providing the dataset used in this study. The authors declare no conflicts of interest.

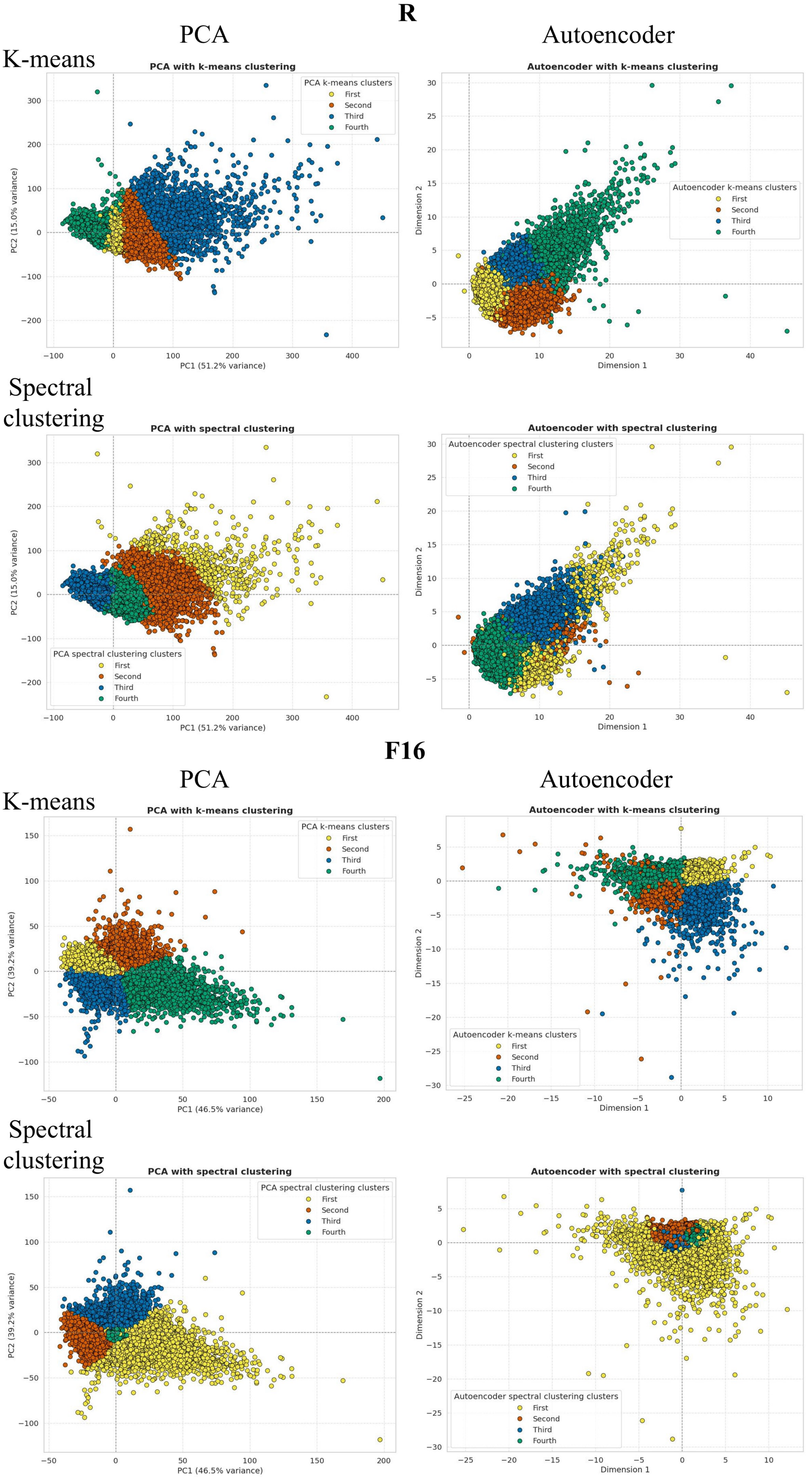

R
PCA
Autoencoder
K-means
PCA with k-means clustering
PCA k-means clusters
First
Second
Third
Fourth
PC1 (51.2% variance)
PC2 (15.0% variance)
Autoencoder with k-means clustering
Autoencoder k-means clusters
First
Second
Third
Fourth
Dimension 1
Dimension 2
Spectral clustering
PCA with spectral clustering
PCA spectral clustering clusters
First
Second
Third
Fourth
PC1 (51.2% variance)
PC2 (15.0% variance)
Autoencoder with spectral clustering
Autoencoder spectral clustering clusters
First
Second
Third
Fourth
Dimension 1
Dimension 2
F16
PCA
Autoencoder
K-means
PCA with k-means clustering
PCA k-means clusters
First
Second
Third
Fourth
PC1 (46.5% variance)
PC2 (39.2% variance)
Autoencoder with k-means clsutering
Autoencoder k-means clusters
First
Second
Third
Fourth
Dimension 1
Dimension 2
Spectral clustering
PCA with spectral clustering
PCA spectral clustering clusters
First
Second
Third
Fourth
PC1 (46.5% variance)
PC2 (39.2% variance)
Autoencoder with spectral clustering
Autoencoder spectral clustering clusters
First
Second
Third
Fourth
Dimension 1
Dimension 2

**Figure 1.** Clusters of individual cows identified by the developed clustering approaches, visualized using the first two principal components (PC1–PC2) from principal component analysis (PCA) and the two highest-variance latent dimensions from the autoencoder (ordered by decreasing variance). Pre-processing using R = all spectra or F16 = only 212 informative wavenumbers from Grelet et al. (2016).

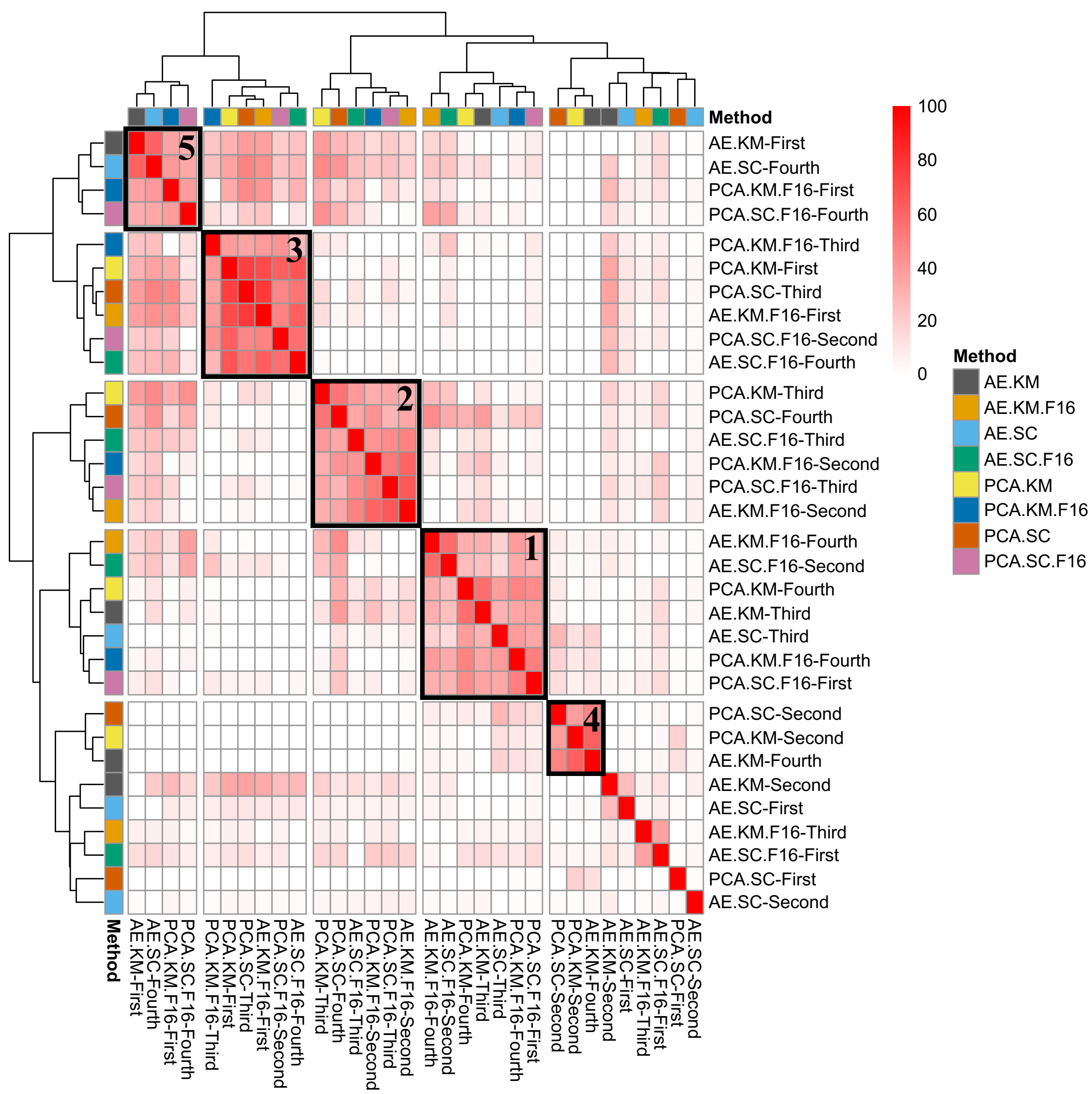

**Figure 2.** Heatmap of Jaccard similarities between cluster assignments. Each cell shows the Jaccard index between a pair of clusters (white = low, red = high); deeper red indicates greater overlap between cluster assignments. Clusters come from eight approaches: AE.KM (R), AE.KM.F16, AE.SC (R), AE.SC.F16, PCA.KM (R), PCA.KM.F16, PCA.SC (R), and PCA.SC.F16, where R = all spectra; F16 = Grelet et al. (2016) wavenumbers; AE = autoencoder; PCA = principal component analysis; KM = k-means; SC = spectral clustering. With each approach, cluster numbers—first, second, third, and fourth—are shown. Black boxes delineate five meta-clusters formed by grouping highly similar cluster assignments.

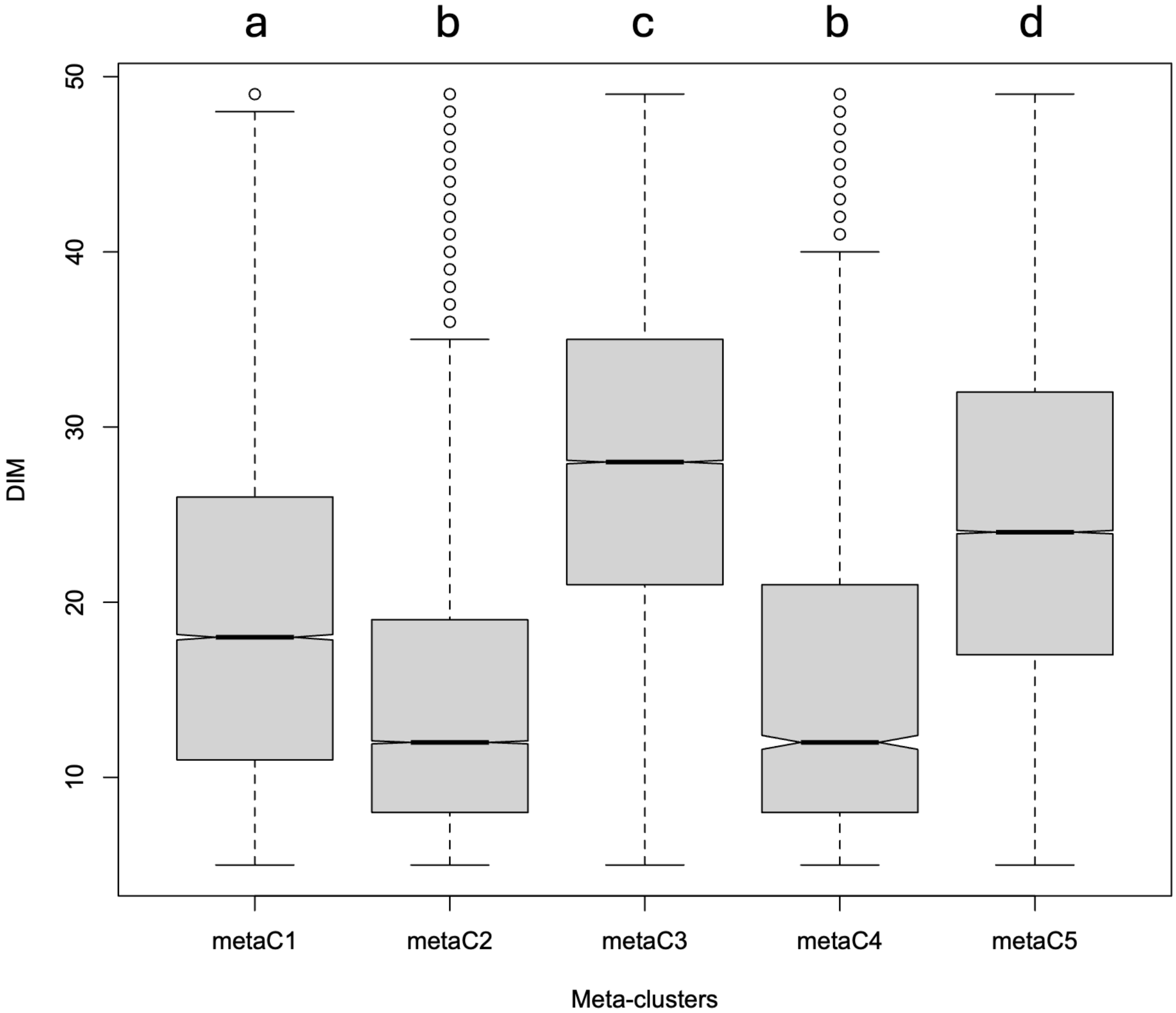

**Figure 3.** Distribution of days in milk (DIM) across meta-clusters. The five meta-clusters of cows are: metaC1 (n = 22,773; 12.84%), metaC2 (n = 40,216; 22.68%), metaC3 (n = 51,117; 28.83%), metaC4 (n = 2,639; 1.49%), and metaC5 (n = 60,577; 34.16%). Letters above the boxplots (a, b, c, d) indicate significant differences among groups. Boxplots that do not share the same letter are significantly different ($P < 0.05$).

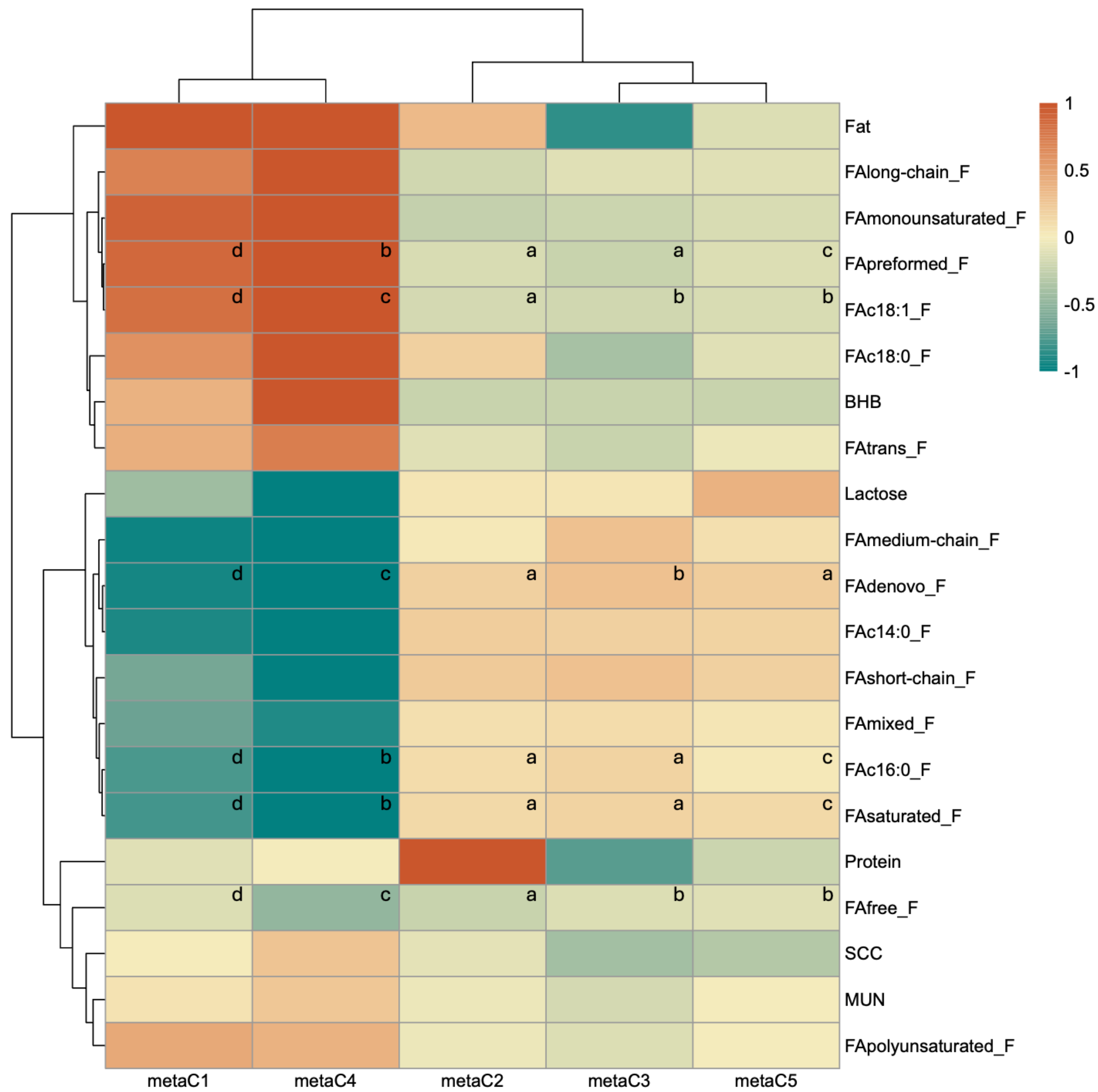

**Figure 4.** Median values of standardized milk traits across the five identified meta-clusters. Fatty acids (FA) are expressed on a fat basis. Abbreviations: BHB, β-hydroxybutyrate; SCC, somatic cell count; MUN, milk urea nitrogen. The five meta-clusters are metaC1, metaC2, metaC3, metaC4, and metaC5. Color scale: deeper red indicates higher trait values; deeper blue indicates lower trait values. Letters above the boxplots (a, b, c, d) indicate significant differences among groups. Boxplots that do not share the same letter are significantly different ($P < 0.05$). Traits without letters were significantly different across all meta-clusters. Letters were omitted to improve figure readability.